\documentclass{article}
\usepackage{comment}

\usepackage{microtype}
\usepackage{graphicx}
\usepackage{subcaption}
\usepackage{booktabs} 

\usepackage{hyperref}

\usepackage[preprint]{icml2026}

\usepackage{amsmath}
\usepackage{amssymb}
\usepackage{mathtools}
\usepackage{amsthm}

\usepackage[capitalize,noabbrev]{cleveref}

\theoremstyle{plain}
\newtheorem{theorem}{Theorem}[section]
\newtheorem{proposition}[theorem]{Proposition}
\newtheorem{lemma}[theorem]{Lemma}
\newtheorem{corollary}[theorem]{Corollary}
\theoremstyle{definition}

\theoremstyle{remark}

\usepackage[textsize=tiny]{todonotes}

\icmltitlerunning{ASEHybrid: Geometry-Aware Graph Neural Networks}

\begin{document}

\twocolumn[
  \icmltitle{ASEHybrid: When Geometry Matters Beyond Homophily\\
in Graph Neural Networks}



\begin{icmlauthorlist}
  \icmlauthor{Shalima Binta Manir}{umbc}
  \icmlauthor{Tim Oates}{umbc}
\end{icmlauthorlist}

\icmlaffiliation{umbc}{Department of Computer Science, University of Maryland, Baltimore County, USA}

\icmlcorrespondingauthor{Shalima Binta Manir}{smanir1@umbc.edu}
\icmlcorrespondingauthor{Tim Oates}{oates@cs.umbc.edu}


  \vskip 0.3in
]



\printAffiliationsAndNotice{}  

\begin{abstract}
Standard message-passing graph neural networks (GNNs) often struggle on graphs with low homophily, yet homophily alone does not explain this behavior, as graphs with similar homophily levels can exhibit markedly different performance and some heterophilous graphs remain easy for vanilla GCNs. Recent work suggests that \emph{label informativeness} (LI), the mutual information between labels of adjacent nodes, provides a more faithful characterization of when graph structure is useful. In this work, we develop a unified theoretical framework that connects curvature-guided rewiring and positional geometry through the lens of label informativeness, and instantiate it in a practical geometry-aware architecture, \emph{ASEHybrid}. Our analysis provides a \emph{necessary-and-sufficient} characterization of when geometry-aware GNNs can improve over feature-only baselines: such gains are possible if and only if graph structure carries label-relevant information beyond node features. Theoretically, we relate adjusted homophily and label informativeness to the spectral behavior of label signals under Laplacian smoothing, show that degree-based Forman curvature does not increase expressivity beyond the one-dimensional Weisfeiler--Lehman test but instead reshapes information flow, and establish convergence and Lipschitz stability guarantees for a curvature-guided rewiring process. Empirically, we instantiate ASEHybrid using Forman curvature and Laplacian positional encodings and conduct controlled ablations on \textsc{Chameleon}, \textsc{Squirrel}, \textsc{Texas}, \textsc{Tolokers}, and \textsc{Minesweeper}, observing gains precisely on label-informative heterophilous benchmarks where graph structure provides label-relevant information beyond node features, and no meaningful improvement in high-baseline regimes.
\end{abstract}


\section{Introduction}

Graph neural networks (GNNs) implement a message-passing paradigm in which node representations are iteratively updated by aggregating information from their neighbors \cite{scarselli2008graph,gilmer2017neural}.
This mechanism underlies many successful architectures and performs particularly well on graphs with \emph{homophilous} structure, where adjacent nodes tend to share labels \cite{kipf2017semi}.
However, on \emph{heterophilous} graphs where neighbors often belong to different classes—standard GNNs can fail dramatically, and in some cases simple feature-only classifiers match or outperform graph-based models \cite{zhu2020beyond,platonov2023characterizing}.

The degree of homophily by itself is insufficient to characterize when message passing helps. Different homophily definitions can disagree sharply and are not directly comparable across datasets with varying class counts or imbalance \cite{zhu2020beyond}.
Moreover, there exist heterophilous graphs on which vanilla GCNs perform well, as well as graphs with similar homophily levels but markedly different GNN performance \cite{ma2022is}.
To address these limitations, recent work has proposed \emph{adjusted homophily} and \emph{label informativeness} (LI) as more principled structural descriptors \cite{platonov2023characterizing}.
LI captures how informative a neighbor’s label is about a node’s label, and empirically correlates more strongly with GNN performance than homophily alone. Platonov et al. \cite{platonov2023characterizing} view label informativeness as an intrinsic property of the data distribution.
We build on this insight by adopting a model-level perspective, characterizing when graph structure
can be exploited by learning architectures via the conditional mutual information \( I(Y;E \mid X) \), which quantifies how much label-relevant information about \(Y\) the graph structure \(E\) contains beyond node features \(X\),
and analyzing how curvature-based signals, positional encodings, and rewiring operationalize (or fail
to operationalize) this information.
Our model is designed for settings in which graph structure provides label-relevant information beyond node features, rather than for feature-dominated regimes. Since homophily alone does not determine whether edges are informative in this sense, our analysis and experiments focus on datasets where label informativeness, rather than homophily, is the primary explanatory factor.

In parallel, there has been growing interest in \emph{geometric augmentations} for GNNs, including structural encodings, positional encodings, and graph rewiring.
Local Curvature Profiles (LCP), based on discrete Ricci curvature, provide node-level summaries of local graph geometry and have been shown to improve performance, especially when combined with Laplacian positional encodings (LapPE) \cite{fesser2024effective,dwivedi2020generalization}.
Curvature-guided rewiring modifies the graph structure by adding edges to mitigate over-squashing \cite{topping2022understanding}.

While recent work has explored combining structural and positional encodings 
\cite{fesser2024effective}, integrating curvature-based encodings with curvature-guided 
rewiring to jointly influence both representations and topology remains unexplored.
In contrast, our proposed model, \emph{ASEHybrid}, integrates node-level structural
encodings (LCP), global positional encodings (LapPE), edge-level curvature conditioning,
and curvature-guided rewiring within a single architecture, allowing geometric
information to jointly shape feature representations and message passing.

This variability raises several fundamental questions:
\begin{itemize}
    \item \textbf{When should curvature and positional geometry help?}
    \item \textbf{What is the role of homophily and label informativeness?}
    \item \textbf{Can curvature-guided rewiring be formalized?}
\end{itemize}

In this work, we address these questions by developing a unified theoretical framework that connects curvature-guided rewiring and positional geometry from the perspective of label informativeness, and by grounding the analysis in carefully controlled experiments on heterophilous benchmarks.

\subsection{Our Contributions}

\paragraph{(A) Structural characterization via adjusted homophily and label informativeness}
We adopt adjusted homophily and label informativeness (LI) as the primary structural descriptors of node-classification graphs \cite{platonov2023characterizing}.
Within this framework:
\begin{itemize}
    \item We relate adjusted homophily to the spectral concentration of label signals under
    Laplacian smoothing, explaining why standard GCNs are biased toward low-frequency,
    homophilous label structure.
   \item We show that LI naturally arises in an information-theoretic comparison between models that can exploit graph edges and those restricted to node features (and positional encodings), yielding a condition of the form $I(Y; E \mid X) > 0$ under which access to edges strictly reduces Bayes risk.
\end{itemize}
\paragraph{(B) Curvature-weighted diffusion and expressivity}
We formalize Forman curvature as a local, degree-based edge attribute \cite{forman2003bochner,sreejith2016forman} and analyze its role in message passing:
\begin{itemize}
    \item We show that incorporating Forman curvature as a local edge feature does \emph{not}
    increase expressive power beyond the one-dimensional Weisfeiler–Lehman (1-WL) test, as
    standard message-passing GNNs are at most as powerful as 1-WL in distinguishing graph
    structures \cite{xu2019powerful}.
    \item In contrast, we show that non-local edge attributes and sufficiently rich positional encodings can strictly increase the distinguishing power of GNNs over plain 1-WL.
\end{itemize}

\paragraph{(C) Curvature-guided rewiring: convergence and stability}
We study a simple curvature-guided rewiring process inspired by discrete geometric flows \cite{topping2022understanding}.
We prove finite termination and convergence under a Lyapunov-style objective based on curvature regularization, and establish Lipschitz stability with respect to bounded edge-edit perturbations.

\begin{figure}[ht]
  \vskip 0.2in
  \begin{center}
   \centerline{\includegraphics[width=\columnwidth]{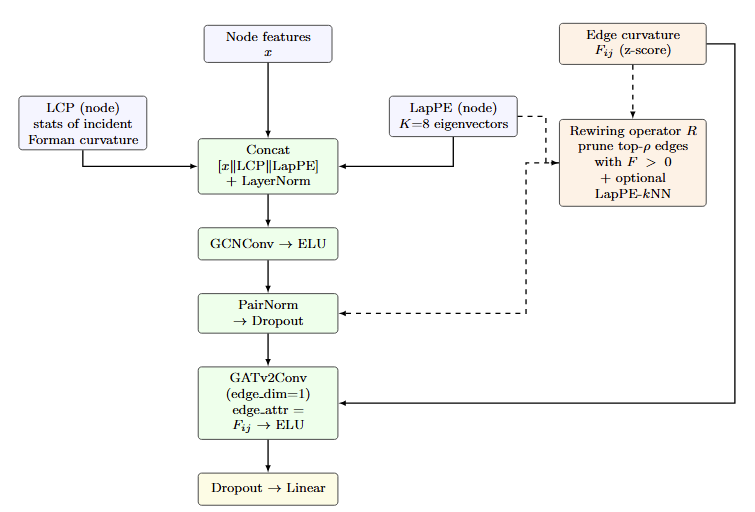}}
    \caption{
      \textbf{ASEHybrid overview.}
      Node-side inputs (features $x$, LCP, LapPE) are concatenated and processed
      sequentially by GCNConv and curvature-conditioned GATv2 attention
      (edge\_attr $=F_{ij}$), followed by a linear classifier.
      Curvature-guided rewiring $R$ is a separate operator that edits the graph
      topology and is enabled only in rewiring-based variants
      (\emph{+Rewire} and \emph{+Both}).
    }
    \label{fig:asehybrid-overview}
  \end{center}
\end{figure}

\paragraph{(D) ASEHybrid: a practical geometry-aware architecture}
We instantiate the framework in \emph{ASEHybrid}, a geometry-aware architecture that combines node features, Forman curvature summaries (LCP-style statistics), and Laplacian positional encodings \cite{dwivedi2022lspe}, followed by curvature-aware attention.  Figure~\ref{fig:asehybrid-overview} provides an overview of ASEHybrid.

\paragraph{(E) Empirical validation and limitations}
Through controlled ablations on the \textsc{Chameleon}, \textsc{Squirrel}, \textsc{Texas}, \textsc{Tolokers}, and \textsc{Minesweeper} datasets, we show that geometry-aware GNNs yield large gains precisely when graph structure is label-informative, and offer negligible benefit otherwise. 
These results empirically validate our theoretical characterization of when geometric augmentation can improve over feature-only and position-only baselines, and complement prior work that identified label informativeness as a descriptive dataset property.

\section{Background and Problem Setup}

\subsection{Graphs and Message-Passing GNNs}

We consider node classification on an undirected simple graph
$G=(V,E)$, where $V$ is the set of nodes, $E \subseteq V \times V$ the edge set,
$x_v \in \mathbb{R}^d$ denotes the feature vector of node $v$, and
$y_v \in \{1,\dots,C\}$ its class label.
For a node $v \in V$, we write $\mathcal{N}(v)$ for its set of neighbors and
$d_v := |\mathcal{N}(v)|$ for its degree.
We use message-passing GNNs, where node representations are updated by
aggregating information from neighboring nodes (formal definition in
Appendix~\ref{app:mpnn}). Standard message-passing GNNs are known to have limited expressive power;
we revisit this issue in Section~4.
In this work we focus on GATv2-style attention mechanisms.
A standard GATv2 layer computes
\[
h_v^{(\ell+1)} =
\sigma\!\left(
\sum_{u \in \mathcal{N}(v)}
\alpha_{uv}^{(\ell)} W h_u^{(\ell)}
\right),
\]
where attention coefficients $\alpha_{uv}^{(\ell)}$ depend on the embeddings
$h_u^{(\ell)}$ and $h_v^{(\ell)}$.
In geometry-aware variants, the attention may additionally condition on a
scalar edge attribute $e_{uv}$, yielding
\[
\alpha_{uv}^{(\ell)} = \mathrm{Attn}
\!\left(h_u^{(\ell)}, h_v^{(\ell)}, e_{uv}\right).
\]

This formulation will allow us to incorporate curvature-based edge signals
while remaining within the message-passing framework.

\subsection{Homophily, Adjusted Homophily, and Label Informativeness}

A classical lens for analyzing GNN performance is \emph{homophily}, the tendency
of edges to connect nodes with the same label.
A commonly used edge homophily measure is
\[
\begin{aligned}
h_{\mathrm{edge}} 
&:= \Pr(y_u = y_v \mid (u,v) \in E) \\
&= \frac{|\{(u,v)\in E : y_u = y_v\}|}{|E|}.
\end{aligned}
\]

However, raw homophily values are difficult to compare across datasets:
they depend strongly on the number of classes and on class imbalance.

To address this issue, we adopt the \emph{adjusted homophily} (assortativity)
coefficient.
Let
\[
\bar p(k) := \frac{1}{2|E|} \sum_{v : y_v = k} d_v
\]
denote the degree-weighted label distribution.
The adjusted homophily is defined as
\begin{equation}
h_{\mathrm{adj}} :=
\frac{
h_{\mathrm{edge}} - \sum_{k=1}^C \bar p(k)^2
}{
1 - \sum_{k=1}^C \bar p(k)^2
}.
\label{eq:adjusted-homophily}
\end{equation}
This quantity equals $1$ for perfectly homophilous graphs,
$0$ for graphs whose edges are independent of labels under a
configuration-model null, and is negative for systematically heterophilous
graphs.

While adjusted homophily captures whether a graph is globally homo- or
heterophilous, it does not directly quantify how informative neighborhoods
are for prediction.
To this end, we use \emph{label informativeness} (LI).
Let $(\xi,\eta)$ be an ordered pair of endpoints of a uniformly sampled edge,
and let $y_\xi, y_\eta$ be their labels.
Label informativeness is defined as the normalized mutual information
\begin{equation}
\mathrm{LI} :=
\frac{I(y_\xi; y_\eta)}{H(y_\xi)}
=
\frac{H(y_\xi) - H(y_\xi \mid y_\eta)}{H(y_\xi)} \in [0,1].
\label{eq:li}
\end{equation}

LI equals $0$ when neighbor labels are statistically independent and equals
$1$ when a neighbor’s label uniquely determines the node’s label.
Empirically and theoretically, LI correlates much more strongly with GNN
performance than homophily alone.
In particular, there exist graphs that are heterophilous but informative
(low $h_{\mathrm{adj}}$ and high LI), as well as heterophilous and uninformative
(low $h_{\mathrm{adj}}$ and low LI).

In our framework, $h_{\mathrm{adj}}$ characterizes the global label structure,
while LI quantifies the potential usefulness of local neighborhoods.
ASEHybrid is designed to be sensitive to the latter: its geometric encodings
and rewiring mechanisms aim to increase the \emph{effective} label
informativeness of the training graph.

\subsection{When Can Graph Structure Help?}

The distinction between homophily and label informativeness naturally leads
to a more fundamental question: \emph{when can access to graph structure
improve prediction over feature-only models?}

Let $X = (x_v)_{v\in V}$ denote node features, $E$ the edge set, and
$Y = (y_v)_{v\in V}$ the labels.
In general, a baseline predictor may have access not only to $X$ but also to
additional node-level information that does not explicitly use $E$ at test time
(e.g., fixed positional encodings).
We denote by $\tilde X$ the total node-side information available to such a
baseline (e.g., $\tilde X=X$ or $\tilde X=[X,\mathrm{LapPE}]$).

We consider the conditional mutual information
\[
I(Y; E \mid \tilde X),
\]
which measures how much additional information about the labels is carried
by the graph structure beyond what is already contained in $\tilde X$.

If $I(Y; E \mid \tilde X) = 0$, then the labels are conditionally independent
of the edges given $\tilde X$, and access to graph structure cannot improve the
Bayes-optimal prediction risk over predictors that only use $\tilde X$.
Conversely, if $I(Y; E \mid \tilde X) > 0$, then there exist data distributions
for which models that can exploit edge-derived signals achieve strictly lower
risk than any predictor based only on $\tilde X$.

This information-theoretic criterion provides a unifying lens for the methods
studied in this paper.
Curvature-based encodings, positional geometry, and curvature-guided rewiring
are useful only insofar as they extract or amplify label-relevant information
from $E$ that is not already present in $\tilde X$.
We formalize this principle in Section~\ref{sec:when-geometry-helps}.

\section{Geometry-Aware Message Passing}

\subsection{Limitations of Laplacian Message Passing}

Many popular GNN architectures, including GCNs and related models, implement
message passing through variants of Laplacian smoothing.
Repeated Laplacian-based propagation in GCN-style models admits a spectral
characterization in the eigenbasis of the normalized Laplacian, where each
component is scaled by a depth-dependent attenuation factor.
A full derivation is provided in Appendix~\ref{app:spectral-laplacian}.

\subsection{Curvature-Weighted Diffusion}
The formal definition of curvature-weighted diffusion and the corresponding
message-passing update are provided in Appendix~\ref{app:curvature-diffusion}.

\begin{proposition}[Curvature-weighted diffusion reduces cross-class mixing]
\label{prop:curvature-mixing}
Let $G=(V,E)$ be a graph with node labels $y:V\to\{1,\dots,C\}$.
Define the baseline GCN transition kernel
\[
p_{ij} = \frac{w_{ij}}{\sum_k w_{ik}},
\qquad
w_{ij} := \frac{\mathbf{1}\{(i,j)\in E\}}{\sqrt{d_i d_j}},
\]
and a curvature-weighted kernel
\[
q_{ij} = \frac{w_{ij}s_{ij}}{\sum_k w_{ik}s_{ik}},
\]
where $s_{ij}\ge 0$ is a curvature-dependent score.

Suppose that for each node $i$, if $J\sim p_{i\cdot}$ and we define
\[
D := \mathbf{1}\{y_i\neq y_J\},
\qquad
S := s_{iJ},
\]
then $\operatorname{Cov}(S,D)\le 0$.
Then the expected cross-class mixing satisfies
\[
\frac{1}{|V|}\sum_{i,j} q_{ij}\mathbf{1}\{y_i\neq y_j\}
\;\le\;
\frac{1}{|V|}\sum_{i,j} p_{ij}\mathbf{1}\{y_i\neq y_j\}.
\]
\end{proposition}
\paragraph{Proof.}
Deferred to Appendix~\ref{app:proof-curvature-mixing}.

\begin{lemma}[Sufficient condition via monotonicity]
\label{lem:monotonicity}
If $S=f(U)$ for a non-increasing function $f$, and
$\mathbb{E}[D\mid U]$ is non-decreasing in $U$, then
$\operatorname{Cov}(S,D)\le 0$.
\end{lemma}
\paragraph{Proof.}
Deferred to Appendix~\ref{app:proof-monotonicity}.

\subsection{Spectral Positional Geometry}

Curvature captures local geometric structure but does not encode global
positional information.
To incorporate global geometry, we use Laplacian positional encodings (LapPE).

Let $L=I-\tilde A$ be the normalized Laplacian with eigenpairs
$(\lambda_k,\phi_k)$.
Define the positional encoding matrix
\[
P=[\phi_2,\dots,\phi_{K+1}]\in\mathbb{R}^{|V|\times K},
\]
excluding the trivial constant eigenvector.
Appending $P$ to node features provides each node with coordinates in a global
spectral embedding that are not smoothed by message passing.

\begin{proposition}[Laplacian positional encodings mitigate spectral bias]
\label{prop:lappe-bias}
Let $\mathcal{H}^{(T)}_{\mathrm{GCN}}$ denote the hypothesis class of a $T$-layer
linearized GCN without positional encodings, and
$\mathcal{H}^{(T)}_{\mathrm{LapPE}}$ the corresponding class augmented with the
first $K$ nontrivial Laplacian eigenvectors.
Then for any label signal $y$,
\[
\min_{\hat y\in\mathcal{H}^{(T)}_{\mathrm{LapPE}}}
\|y-\hat y\|_2
\;\le\;
\|y-\Pi_{\mathrm{span}(\phi_2,\dots,\phi_{K+1})}y\|_2,
\]
provided $(1-\lambda_k)^T\neq 0$ for all $k=2,\dots,K+1$.
\end{proposition}

\paragraph{Interpretation.}
The right-hand side of the inequality is the energy of the label signal $y$
outside the span of the injected Laplacian eigenvectors.
The proposition therefore states that a $T$-layer GCN augmented with Laplacian
positional encodings can approximate $y$ at least as well as its projection
onto this spectral subspace.
In particular, spectral components that would be attenuated by repeated
Laplacian smoothing are preserved when supplied as positional inputs, provided
they are not annihilated by propagation.

\paragraph{Proof.}
Deferred to Appendix~\ref{prop:lappe-bias}.

Curvature-weighted diffusion and Laplacian positional encodings play
complementary roles.
Curvature reshapes local aggregation to suppress harmful mixing, while LapPE
injects global, high-frequency structure that standard message passing cannot
recover.
Neither mechanism increases expressive power by itself; rather, they reshape
information flow, a distinction formalized in the next section.
\section{Expressivity of Geometry-Aware GNNs}

In this section, we analyze the expressive power of geometry-aware
message-passing GNNs.
We distinguish between (i) local, degree-based geometric signals such as
Forman curvature, (ii) non-local edge attributes derived from larger graph
motifs, and (iii) positional encodings that provide global symmetry-breaking
information.
Our results clarify which forms of geometric augmentation can—and cannot—
increase expressive power beyond the one-dimensional Weisfeiler--Lehman (1-WL)
test.

\subsection{No Expressivity Gain from Local Curvature}

We first show that incorporating degree-based Forman curvature as an edge
attribute does not increase the expressive power of message-passing GNNs.

Recall that the Forman curvature of an undirected edge $(u,v)$ is
\[
F(u,v) = 4 - \deg(u) - \deg(v),
\]
which depends only on the degrees of its endpoints.

\begin{theorem}[No WL gain from degree-based Forman curvature]
\label{thm:no-wl-curvature}
Any message-passing GNN that uses degree-based Forman curvature as an edge
feature is no more expressive than the one-dimensional Weisfeiler--Lehman
(1-WL) test.
\end{theorem}
\paragraph{Proof.}
Deferred to Appendix~\ref{app:proof-no-wl-curvature}.
\paragraph{Implication.}
Local curvature does not act as a symmetry-breaking signal.
Any empirical gains from curvature must therefore arise from improved
information flow or stability, rather than increased representational
capacity.

\subsection{Gains from Non-Local Edge Attributes}

We next contrast local curvature with non-local edge attributes that encode
information beyond node degrees.

\begin{theorem}[Expressivity gain with non-local edge attributes]
\label{thm:nonlocal-edge}
There exist pairs of non-isomorphic graphs that are indistinguishable by
1-WL, but can be distinguished by message-passing GNNs equipped with
appropriate non-local, isomorphism-invariant edge attributes.
\end{theorem}
\paragraph{Proof.}
Deferred to Appendix~\ref{app:proof-nonlocal-edge}.
\subsection{Positional Encodings as Symmetry Breakers}

We now turn to positional encodings, which provide nodes with global,
graph-dependent coordinates.

\begin{theorem}[PE-augmented expressivity gain]
\label{thm:pe-expressivity}
There exist graph pairs that are indistinguishable by 1-WL, but can be
distinguished with high probability by message-passing GNNs augmented with
sufficiently rich positional encodings.
\end{theorem}
\paragraph{Proof.}
Deferred to Appendix~\ref{app:proof-pe-expressivity}.

\paragraph{Summary.}
Local curvature-based edge features do not increase expressive power beyond
1-WL, while non-local edge attributes and positional encodings can.
This separation explains why curvature improves performance through improved
diffusion and stability rather than symmetry breaking, and why combining
curvature with positional geometry as in ASEHybrid is essential.

\section{Curvature-Guided Rewiring}

We study a curvature-guided rewiring process that modifies the graph structure
during training. The goal is to remove edges that are geometrically redundant
under a curvature-based score and to add edges that are consistent with global
positional geometry (e.g., LapPE). We formalize the rewiring map and analyze
its convergence and stability.
\paragraph{Rewiring operator.}
The rewiring procedure is formally defined in
Appendix~\ref{app:rewiring-algorithm}.
\subsection{Convergence of the Rewiring Dynamics}

\begin{proposition}[Finite termination of rewiring]
\label{prop:finite-termination}
Suppose there exists a scalar objective $\mathcal{L}(E)$ such that
$\mathcal{L}(E_{t+1}) < \mathcal{L}(E_t)$ whenever $E_{t+1}\neq E_t$.
If the admissible edge sets form a finite space, then the rewiring process
terminates after finitely many steps.
\end{proposition}
\paragraph{Proof.}
Deferred to Appendix~\ref{app:proof-finite-termination}.
\subsection{Convergence and Local Stability of the Rewiring Map}

We now show that, under eigengap and margin assumptions, the rewiring operator
is locally constant around a fixed point.

\begin{theorem}[Local stabilization of the rewiring map]
\label{thm:local-stabilization}
Let $E^\star$ be a fixed point of $\mathcal{R}$.
Assume:
\begin{enumerate}
    \item \textbf{Eigengap.} The normalized Laplacian of $G^\star=(V,E^\star)$
    has an eigengap separating the retained LapPE subspace from the rest of the
    spectrum.

    \item \textbf{$k$NN margin.} For each node $u$, the $k$-th and $(k+1)$-th
    nearest-neighbor distances in LapPE space are separated by a positive
    margin at $E^\star$.

    \item \textbf{Pruning-score margin.}
    Let $(e_1^\star,\dots,e_{|E^{\star+}|}^\star)$ be the ordering of
    $E^{\star+}:=\{e\in E^\star:s^\star(e)>0\}$ by decreasing scores. Then
    \[
    s^\star(e_{\lceil\rho |E^{\star+}|\rceil}^\star)
    >
    s^\star(e_{\lceil\rho |E^{\star+}|\rceil+1}^\star)
    \]
    (strict separation at the pruning threshold).

    \item \textbf{Score stability.} The score map $E\mapsto s_E(\cdot)$ is
    continuous (or Lipschitz) in a neighborhood of $E^\star$ under bounded
    edge edits.
\end{enumerate}
Then there exists a neighborhood $\mathcal{U}$ of $E^\star$ such that for all
$E\in\mathcal{U}$,
\[
\mathcal{R}(E)=\mathcal{R}(E^\star)=E^\star.
\]
In particular, $\mathcal{R}$ is locally constant (and hence locally stable)
at $E^\star$.
\end{theorem}
\paragraph{Proof.}
Deferred to Appendix~\ref{app:proof-local-stabilization}.

\subsection{Stability Under Edge Edits}

We now bound how much the normalized adjacency changes under bounded edge
edits, and propagate this to embedding stability.

Let $\tilde A := D^{-1/2}(A+I)D^{-1/2}$ denote the self-loop augmented
normalized adjacency, and let $\tilde d_{\min}\ge 1$ denote the minimum
degree after adding self-loops.

\begin{lemma}[Operator-norm perturbation under edge edits]
\label{lem:operator-perturbation}
Let $E$ and $E'$ be edge sets that differ by at most $K$ edges.
Let $\tilde A(E)$ and $\tilde A(E')$ denote the corresponding self-loop
normalized adjacency matrices.
Assume the degree ratio is bounded, i.e.,
$\tilde d_{\max}/\tilde d_{\min} \le c$ for some constant $c$.
Then
\[
\|\tilde A(E')-\tilde A(E)\|_2 \;\le\; C\,\frac{K}{\tilde d_{\min}},
\]
for a constant $C$ depending only on $c$.
\end{lemma}

\paragraph{Proof.}
Deferred to Appendix~\ref{app:proof-operator-perturbation}.

\begin{corollary}[Lipschitz stability under bounded rewiring]
\label{cor:lipschitz-stability}
Consider a $T$-layer message-passing network with propagation operator
$\tilde A(E)$ at each layer and weight matrices $W_1,\dots,W_T$.
Assume the pointwise nonlinearity is 1-Lipschitz and
$\tilde d_{\max}/\tilde d_{\min}\le c$.
If $E$ and $E'$ differ by at most $K$ edges, then the output embeddings satisfy
\[
\|H_T(E')-H_T(E)\|_2
\;\le\;
C\,\frac{K}{\tilde d_{\min}}
\prod_{\ell=1}^T \|W_\ell\|_2 \,\|H_0\|_2 ,
\]
where $C$ depends only on $c$.
\end{corollary}

\paragraph{Proof.}
Deferred to Appendix~\ref{app:proof-lipschitz-stability}.

Curvature-guided rewiring defines a discrete geometric flow over graphs.
Under a strictly decreasing objective it terminates finitely, under eigengap
and margin assumptions it stabilizes locally, and under bounded edge edits it
induces controlled changes in normalized propagation and thus in embeddings.

\section{When Does Geometry Help?}
\label{sec:when-geometry-helps}

The preceding sections analyzed how curvature-based diffusion, positional
geometry, and rewiring affect information flow and expressive power.
We now address a more fundamental question:
\emph{when can access to graph structure improve prediction over models that
do not explicitly use edges at test time?}

Our answer is information-theoretic.
We show that the usefulness of geometry-aware GNNs is governed by whether
graph structure carries label-relevant information beyond the information
already available to the baseline predictor.

\subsection{Edge--Label Information and Bayes Risk}

Let $X=(x_v)_{v\in V}$ denote node features, $E$ the edge set, and
$Y=(y_v)_{v\in V}$ the node labels.
In general, a baseline model may have access not only to $X$ but also to
additional node-level information that does not adapt to edges at test time
(e.g., fixed positional encodings).
We denote by $\tilde X$ the total node-side information available to such a
baseline.

We consider a probabilistic data-generating process over $(\tilde X,E,Y)$.

A \emph{$\tilde X$-only predictor} is any measurable function
$f:\tilde X\to\Delta(\mathcal Y)$.
An \emph{edge-aware predictor} may additionally depend on $E$,
$f:(\tilde X,E)\to\Delta(\mathcal Y)$.
Prediction quality is measured by expected risk under a proper scoring rule;
for concreteness, we focus on log loss.

Under log loss, the Bayes-optimal risk of a predictor using information $Z$ is
\[
R^\ast(Z)
=
\mathbb{E}[-\log p(Y\mid Z)]
=
H(Y\mid Z),
\]
the conditional entropy of $Y$ given $Z$.

Consequently, the gap between the optimal $\tilde X$-only risk and the optimal
edge-aware risk is
\[
\begin{aligned}
R^\ast(\tilde X)-R^\ast(\tilde X,E)
&=
H(Y\mid \tilde X)-H(Y\mid \tilde X,E) \\
&=
I(Y;E\mid \tilde X),
\end{aligned}
\]
the conditional mutual information between labels and edges given $\tilde X$.

This quantity precisely measures the \emph{potential benefit} of exploiting
graph structure beyond the information already encoded in $\tilde X$.

\subsection{Information-Theoretic Necessity and Sufficiency}

We now state the main result of this section, which provides a sharp
characterization of when geometry-aware methods can help.

\begin{proposition}[Edge--label information governs usefulness]
\label{prop:when-geometry-helps}
Let $(\tilde X,E,Y)$ be jointly distributed.
\begin{enumerate}
    \item (\emph{Necessity})
    If $I(Y;E\mid \tilde X)=0$, then no predictor that has access to $E$ can
    achieve strictly lower Bayes risk than the optimal $\tilde X$-only
    predictor.

    \item (\emph{Sufficiency})
    If $I(Y;E\mid \tilde X)>0$, then there exists a data distribution and an
    edge-aware predictor that achieves strictly lower Bayes risk than any
    $\tilde X$-only predictor.
\end{enumerate}
\end{proposition}

\paragraph{Proof.}
Deferred to Appendix~\ref{app:proof-when-geometry-helps}.
\paragraph{Interpretation.}
Proposition~\ref{prop:when-geometry-helps} provides a complete
characterization:
geometry-aware methods are \emph{provably useless} when edges add no
label-relevant information beyond $\tilde X$, and \emph{provably helpful}
when they do.

When edge features or graph structure do not carry label-relevant information beyond node-side inputs, geometry-aware mechanisms cannot improve prediction in principle. In such regimes, feature-based or position-based models are sufficient, and incorporating edge-derived signals is unnecessary. This follows directly from the information-theoretic criterion governing when access to graph structure can reduce Bayes risk.
\subsection{Implications for Geometry-Aware GNNs}
 A detailed discussion of these implications is provided in Appendix~\ref{app:implications}.



\section{Experiments}
We evaluate \emph{ASEHybrid} and controlled ablations thereof on five
node-classification benchmarks selected to span a range of
heterophilous regimes and label informativeness (LI) levels:
\textbf{Chameleon} and \textbf{Squirrel} (Wikipedia web graphs;
Section~\ref{sec:exp1}),
\textbf{Texas} (WebKB citation network; Section~\ref{sec:exp2}),
\textbf{Minesweeper} (high-baseline synthetic graph; Section~\ref{sec:exp3}),
and \textbf{Tolokers} (large-scale crowdsourcing graph;
Section~\ref{sec:tolokers}).
Following prior work, we use standard splits provided by PyTorch
Geometric~\cite{fey2019fast}.
\footnote{
\footnotesize
Chameleon/Squirrel: WikipediaNetwork~\cite{rozemberczki2021multi};
Texas: WebKB~\cite{craven1998learning};
Minesweeper/Tolokers: HeterophilousGraphDataset~\cite{platonov2023characterizing}.
}

These datasets are chosen to test the central theoretical prediction of
this paper: \emph{geometry-aware mechanisms improve performance precisely
when graph structure carries label-relevant information beyond node-side
features}, and provide little or no benefit otherwise.
\paragraph{Architecture}ASEHybrid combines curvature-augmented node features, curvature-aware
attention, and optional curvature-guided rewiring.
Full architectural details are provided in Appendix~\ref{app:architecture}.
\paragraph{Training.}
All models are trained under a common protocol with standard optimization and
early stopping; full training details are provided in
Appendix~\ref{app:training}.
\paragraph{Curvature-guided rewiring.}
Rewiring prunes a fraction $\rho \in \{0.01, 0.02\}$ of edges with the
highest positive Forman curvature
$F(i,j) = 4 - \deg(i) - \deg(j)$,
and optionally adds $k \in \{0,1\}$ $k$NN edges in Laplacian positional
encoding (LapPE) space.
Unless explicitly stated, rewiring is applied as a \emph{one-shot
preprocessing step}, consistent with the theoretical analysis of
finite termination and local stability.
Periodic rewiring is used only for ablation studies on small graphs.

\paragraph{Baselines and ablations.}
We compare against a vanilla two-layer GCN with ReLU activations, trained
using the same splits and optimizer.
Ablations are reported relative to \textbf{ASE base} and isolate the
effects of (i) curvature-aware message passing and
(ii) curvature-guided graph rewiring.

\begin{table*}[t]
\caption{Test accuracy (\%) on heterophilous benchmarks (mean $\pm$ std).
Chameleon and Squirrel use fixed splits (10 seeds);
Texas, Minesweeper, and Tolokers report results over dataset-provided splits. For Tolokers, curvature-guided rewiring is applied as a one-shot preprocessing step.}
\label{tab:main-results}
\centering
\small
\begin{sc}
\begin{tabular}{lccccc}
\toprule
Method & Chameleon & Squirrel & Texas & Minesweeper & Tolokers \\
\midrule
Vanilla GCN
& 39.56 $\pm$ 2.32
& 28.29 $\pm$ 1.08
& 57.57 $\pm$ 6.12
& 80.04 $\pm$ 0.15
& 78.54 $\pm$ 0.33 \\

ASE base
& 48.05 $\pm$ 1.47
& 32.08 $\pm$ 1.00
& 58.92 $\pm$ 5.06
& 80.06 $\pm$ 0.26
& 81.24 $\pm$ 0.47 \\

+EdgeCurv
& 48.05 $\pm$ 1.21
& 33.00 $\pm$ 0.73
& 58.11 $\pm$ 8.08
& 80.07 $\pm$ 0.17
&  81.28 $\pm$ 0.43 \\

+Rewire
& 64.47 $\pm$ 0.98
& 55.29 $\pm$ 0.70
& 59.19 $\pm$ 9.49
& 79.98 $\pm$ 0.34
& 80.32 $\pm$ 0.39 \\

\textbf{+Both (ASEHybrid)}
& \textbf{64.78 $\pm$ 0.77}
& \textbf{56.13 $\pm$ 0.96}
& \textbf{61.35 $\pm$ 6.87}
& \textbf{80.05 $\pm$ 0.16}
& \textbf{81.51 $\pm$ 0.43} \\

\bottomrule
\end{tabular}
\end{sc}
\end{table*}

\subsection{Ablations on Web Graphs}
\label{sec:exp1}

We begin with Chameleon and Squirrel, two canonical heterophilous web
graphs known to exhibit high label informativeness despite low adjusted
homophily.
These datasets provide a setting where the theory predicts that access
to graph structure should be strongly beneficial.

Rewiring yields the dominant performance gains
(+16--17 points on Chameleon and +23--24 points on Squirrel; see
Table~\ref{tab:main-results}),
while curvature-aware message passing alone provides at most modest
improvements.
This behavior directly reflects the theoretical results:
degree-based curvature does not increase expressive power beyond 1-WL,
whereas curvature-guided rewiring can substantially improve information
flow when edges are label-informative.
\subsection{WebKB-Texas}
\label{sec:exp2}

Texas represents a smaller, moderately heterophilous graph with higher
variance across splits and weaker global structural signal.
Here, the theory predicts that geometry-aware methods may provide
\emph{consistent but modest} gains.

ASEHybrid achieves the highest mean accuracy across splits
(see Table~\ref{tab:main-results}), with an absolute gain of $+3.78$ points over a
vanilla GCN.
While smaller than on the web graphs, the improvement is consistent with
the information-theoretic criterion: edges in Texas carry limited but
nonzero label-relevant information beyond node features, leading to
modest but reliable gains.
\subsection{Minesweeper: High-Baseline Regime}
\label{sec:exp3}

Results on the high-baseline Minesweeper benchmark are reported in Appendix~\ref{app:minesweeper}.
\subsection{Tolokers: Large-Scale Graph}
\label{sec:tolokers}
Tolokers is a large, noisy crowdsourcing graph with moderate label
informativeness and substantial degree heterogeneity.
Due to scale considerations, we apply curvature-guided rewiring as a
single preprocessing step.
ASEHybrid yields a consistent improvement over a vanilla GCN
(see Table~\ref{tab:main-results}), while rewiring alone does not suffice.
The modest gains observed here are consistent with theory:
in large, heterogeneous graphs with noisy edges, geometry-aware
mechanisms can improve stability and diffusion without producing
dramatic performance jumps.


\section{Related Work}
\paragraph{Relation to prior work.}
Our work relates to prior studies on homophily and label informativeness,
curvature-based graph learning, positional encodings, and graph rewiring.
Unlike existing approaches, we provide a unified information-theoretic
characterization of when graph structure can improve prediction, and establish
formal convergence and stability guarantees for curvature-guided rewiring.
A detailed discussion of related work is provided in Appendix~\ref{app:related}.


\section{Conclusion}
We studied geometry-aware graph neural networks through a unified theoretical
and empirical lens, clarifying the roles of homophily, label informativeness,
curvature, positional encodings, and graph rewiring in heterophilous node
classification.

On the theory side, we showed that local, degree-based geometric signals such
as Forman curvature do not increase expressive power beyond the
one-dimensional Weisfeiler--Lehman test, but can substantially improve
performance by reshaping diffusion and reducing harmful cross-class mixing.
In contrast, non-local edge attributes and positional encodings can strictly
increase expressive power by breaking graph symmetries.
For curvature-guided rewiring, we established finite termination, local
stabilization, and Lipschitz stability under bounded edge edits.

These results lead to a sharp information-theoretic characterization of when
geometry helps: geometry-aware mechanisms are useful precisely when graph
structure carries label-relevant information beyond node features, as
measured by conditional edge--label mutual information.
Our experiments corroborate this prediction, showing large gains on
label-informative heterophilous graphs, modest improvements on smaller
benchmarks, and comparable performance when node features alone are already
highly predictive.

We instantiated these insights in \emph{ASEHybrid}, a practical geometry-aware
architecture that consistently outperforms vanilla message passing across
diverse heterophilous benchmarks while remaining computationally efficient.

\section*{Impact Statement}

This paper advances the theoretical understanding and design of
geometry-aware graph neural networks.
The methods studied operate on abstract graph structures and are agnostic to
the semantic content of node features or labels, and therefore do not introduce
new sources of bias beyond those present in the underlying data.
While graph-based models may be applied in sensitive domains, any broader
societal or ethical impacts depend on downstream use and deployment, which are
beyond the scope of this work.


\bibliography{example_paper}
\bibliographystyle{icml2026}

\newpage
\appendix
\onecolumn

\subsection{(A) Notation}

Table~\ref{tab:notation} summarizes the main notation used throughout the paper.

\begin{table}[h]
\centering
\caption{Summary of notation used throughout the paper.}
\label{tab:notation}
\small
\begin{tabular}{ll}
\hline
\textbf{Symbol} & \textbf{Description} \\
\hline
$G=(V,E)$ & Undirected graph with node set $V$ and edge set $E$ \\
$n=|V|$ & Number of nodes \\
$m=|E|$ & Number of edges \\
$v,u,i,j$ & Node indices \\
$\mathcal{N}(v)$ & Neighborhood of node $v$ \\
$d_v$ & Degree of node $v$ \\
$A$ & Adjacency matrix of $G$ \\
$D$ & Diagonal degree matrix \\
$\tilde A$ & Normalized adjacency $D^{-1/2}(A+I)D^{-1/2}$ \\
$L$ & Normalized graph Laplacian $I-\tilde A$ \\
$\lambda_k, \phi_k$ & $k$-th eigenvalue and eigenvector of $L$ \\
$T$ & Number of message-passing layers / propagation steps \\
$K$ & Number of Laplacian positional encoding dimensions \\
$P$ & Laplacian positional encoding matrix $[\phi_2,\dots,\phi_{K+1}]$ \\
$x_v \in \mathbb{R}^d$ & Input feature vector of node $v$ \\
$X$ & Matrix of node features $(x_v)_{v\in V}$ \\
$\tilde X$ & Node-side information available to feature-only or PE-only baselines \\
$y_v \in \{1,\dots,C\}$ & Class label of node $v$ \\
$Y$ & Vector of node labels $(y_v)_{v\in V}$ \\
$h_v^{(\ell)}$ & Node representation of $v$ at layer $\ell$ \\
$H^{(\ell)}$ & Matrix of node representations at layer $\ell$ \\
$W^{(\ell)}$ & Trainable weight matrix at layer $\ell$ \\
$\sigma(\cdot)$ & Pointwise nonlinearity \\
$\mathrm{AGG}$ & Permutation-invariant aggregation operator \\
$e_{uv}$ & Edge attribute associated with edge $(u,v)$ \\
$F_{uv}$ & Degree-based Forman curvature of edge $(u,v)$ \\
$f(\cdot)$ & Curvature-based edge weighting function \\
$p_{ij}$ & Baseline GCN transition kernel \\
$q_{ij}$ & Curvature-weighted transition kernel \\
$s_{ij}$ & Curvature-dependent edge score \\
$h_{\mathrm{edge}}$ & Edge homophily \\
$h_{\mathrm{adj}}$ & Adjusted homophily (assortativity) \\
$\mathrm{LI}$ & Label informativeness \\
$I(\cdot;\cdot)$ & Mutual information \\
$I(\cdot;\cdot \mid \cdot)$ & Conditional mutual information \\
$R^\ast(Z)$ & Bayes-optimal risk given information $Z$ \\
$R$ & Curvature-guided rewiring operator \\
$G_t=(V,E_t)$ & Graph at rewiring step $t$ \\
$E_t$ & Edge set at rewiring step $t$ \\
$\rho$ & Fraction of edges pruned during rewiring \\
$k$ & Number of nearest neighbors added in LapPE space \\
$\tilde d_{\min}, \tilde d_{\max}$ & Minimum / maximum degree after self-loop augmentation \\
\hline
\end{tabular}
\end{table}

\section{Additional Background}
\label{app:background}

\subsection{Message-Passing GNNs and Expressivity}
\label{app:mpnn}

A generic message-passing graph neural network (GNN) updates node
representations according to
\[
h_v^{(\ell+1)} =
\sigma\!\left(
\mathrm{AGG}
\left\{
\phi\!\left(h_v^{(\ell)}, h_u^{(\ell)}, e_{uv}\right)
:\; u \in \mathcal{N}(v)
\right\}
\right),
\]
where $h_v^{(\ell)}$ is the representation of node $v$ at layer $\ell$,
$e_{uv}$ is an optional edge feature,
$\phi$ is a learnable message function,
$\mathrm{AGG}$ is a permutation-invariant aggregation operator,
and $\sigma$ is a nonlinearity.
It is well known that such message-passing architectures are limited in
expressive power: when initialized from the same node features, they are
at most as powerful as the one-dimensional Weisfeiler--Lehman (1-WL) test
for distinguishing non-isomorphic graphs.
This limitation applies even when edge features are present, provided these
features are functions of local graph structure and do not encode non-local
or positional information.

\subsection{Spectral View of Laplacian Message Passing}
\label{app:spectral-laplacian}
Let $A$ denote the adjacency matrix of $G$, $D$ the diagonal degree matrix,
and $\tilde A = D^{-1/2}(A+I)D^{-1/2}$ the normalized adjacency.
A linearized GCN layer takes the form
\[
H^{(\ell+1)} = \tilde A H^{(\ell)} W.
\]

Let $L = I - \tilde A$ denote the normalized Laplacian, with eigendecomposition
$L = \Phi \Lambda \Phi^\top$, where
$0=\lambda_1 \le \lambda_2 \le \cdots \le \lambda_n \le 2$.
Repeated propagation yields
\[
H^{(T)} = \tilde A^T H^{(0)} W^{(T)}
       = \Phi (I-\Lambda)^T \Phi^\top H^{(0)} W^{(T)}.
\]

Thus, each eigenmode $\phi_k$ is scaled by $(1-\lambda_k)^T$, where $T$ is the
number of propagation steps (layers) in the linearized model. Since $I-\Lambda$ is diagonal, $(I-\Lambda)^T$ raises each diagonal entry $(1-\lambda_k)$ to the $T$-th power.
High-frequency components (large $\lambda_k$) are exponentially attenuated,
while low-frequency components dominate.
This induces a strong low-pass bias, favoring label signals that vary smoothly
over edges \cite{oono2020graph}.

In heterophilous graphs, class labels often exhibit significant high-frequency
energy.
As a result, Laplacian smoothing mixes features across class boundaries,
leading to over-smoothing and degraded performance.
This motivates geometry-aware modifications that selectively attenuate harmful
aggregation without abandoning message passing entirely.

\subsection{Curvature-Weighted Diffusion (Definition and Update Rule)}
\label{app:curvature-diffusion}

To modulate message passing based on local graph geometry, we introduce a
degree-based Forman curvature for undirected edges.
For an undirected edge $(i,j)\in E$, we define the (degree-based) Forman
curvature
\[
F_{ij} := 4 - d_i - d_j,
\]
where $d_i := |\mathcal{N}(i)|$ denotes the degree of node $i$ \cite{forman2003bochner}.
Edges connecting high-degree nodes or structurally inconsistent regions tend
to have large negative curvature, while edges inside cohesive neighborhoods
exhibit higher curvature.

In geometry-aware message passing, curvature enters as a scalar edge attribute
that conditions aggregation weights.
A curvature-weighted diffusion step takes the form
\[
h_i^{(\ell+1)} =
\sigma\!\left(
\sum_{j \in \mathcal{N}(i)}
f(F_{ij})
\frac{1}{\sqrt{d_i d_j}}
W h_j^{(\ell)}
\right),
\]
where $f:\mathbb{R}\to\mathbb{R}_{\ge 0}$ is a monotone or learnable weighting
function, typically implemented via attention.

This formulation corresponds to anisotropic diffusion: aggregation across
low-curvature edges is attenuated relative to standard Laplacian smoothing.

We now formalize the intuition that curvature-based reweighting reduces
cross-class mixing.

\section{Additional Proofs}
\label{app:proofs}

\subsection{Proof of Proposition~\ref{prop:curvature-mixing}}
\label{app:proof-curvature-mixing}

\begin{proof}
Fix a node $i$.
Sampling $J\sim q_{i\cdot}$ corresponds to sampling $J\sim p_{i\cdot}$ under
importance weighting by the nonnegative scores $S=s_{iJ}$.
In particular, for any function $\varphi$,
\[
\mathbb{E}_{q}[\varphi(J)]
=
\frac{\mathbb{E}_{p}[\varphi(J)S]}{\mathbb{E}_{p}[S]}.
\]

Applying this identity with $\varphi(J)=D$ yields
\[
\mathbb{E}_{q}[D]
=
\frac{\mathbb{E}_{p}[DS]}{\mathbb{E}_{p}[S]}.
\]
Since $\operatorname{Cov}(S,D)\le 0$, we have
$\mathbb{E}_{p}[DS]\le \mathbb{E}_{p}[D]\mathbb{E}_{p}[S]$.
Substituting gives
$\mathbb{E}_{q}[D]\le \mathbb{E}_{p}[D]$.
Averaging over all nodes $i\in V$ yields the stated inequality.
\end{proof}
The covariance condition admits a natural sufficient condition.
\subsection{Proof of Lemma~\ref{lem:monotonicity}}
\label{app:proof-monotonicity}

\begin{proof}
Let $U'$ be an independent copy of $U$, and let $D'$ be the corresponding copy
of $D$.
Using the identity
\[
\operatorname{Cov}(f(U),D)
=
\tfrac12\,\mathbb{E}\!\left[(f(U)-f(U'))(D-D')\right],
\]
we examine the sign of the integrand.
Since $f$ is non-increasing and $\mathbb{E}[D\mid U]$ is non-decreasing,
$(f(U)-f(U'))(D-D')\le 0$ almost surely.
Taking expectations yields $\operatorname{Cov}(S,D)\le 0$.
\end{proof}

\subsection{Proof of Proposition~\ref{prop:lappe-bias}}
\label{app:proof-lappe}

\begin{proof}
Let $L=\Phi\Lambda\Phi^\top$ be the normalized Laplacian eigendecomposition.
Repeated propagation yields
\[
\tilde A^T=\Phi(I-\Lambda)^T\Phi^\top.
\]
On the subspace $\mathrm{span}(\phi_2,\dots,\phi_{K+1})$, the operator
$\tilde A^T$ is invertible under the stated condition.
Thus any signal in this subspace can be exactly represented by propagating
the positional encodings, and the best approximation error over
$\mathcal{H}^{(T)}_{\mathrm{LapPE}}$ is bounded by the projection residual.
\end{proof}

\subsection{Proof of Theorem~\ref{thm:no-wl-curvature}}
\label{app:proof-no-wl-curvature}

\begin{proof}
We prove the claim by a simulation argument.

Let $G$ and $G'$ be two graphs that are indistinguishable by 1-WL.
By definition, there exists a sequence of bijections
$\phi_t:V(G)\to V(G')$ preserving node colors at every WL iteration.

After the first WL iteration, the color of each node uniquely determines its
degree.
Thus, degrees are 1-WL-definable.
Since the Forman curvature $F(u,v)$ is a deterministic function of
$\deg(u)$ and $\deg(v)$, the multiset of curvature values on edges incident to
a node is also preserved under $\phi_t$.

Consider a message-passing GNN that conditions aggregation on $F(u,v)$.
At each layer, the updated node representation is a function of the
multiset of neighbor representations and curvature values.
Because both node degrees and curvature values are invariant under the WL
bijections $\phi_t$, the entire message-passing update can be simulated by
a 1-WL refinement started from the same initial node labels.

Formally, this follows from the fact that initializing 1-WL with labels that
are deterministic functions of previous WL colors does not increase its
distinguishing power.
Therefore, curvature-augmented message passing cannot distinguish any pair
of graphs that 1-WL fails to distinguish, and does not increase expressive
power beyond 1-WL.
\end{proof}

\subsection{Proof of Theorem~\ref{thm:nonlocal-edge}}
\label{app:proof-nonlocal-edge}

\begin{proof}
Consider two $d$-regular graphs $G_1$ and $G_2$ on the same number of nodes,
where $G_1$ is triangle-free and $G_2$ contains at least one triangle.
Such graphs are indistinguishable by 1-WL.

Define an edge attribute
\[
e(u,v) := |\mathcal{N}(u)\cap\mathcal{N}(v)|,
\]
the number of common neighbors of $u$ and $v$.
This attribute is invariant under graph isomorphisms but is not determined
by node degrees alone.

In $G_1$, all edges satisfy $e(u,v)=0$, whereas in $G_2$, edges participating
in triangles satisfy $e(u,v)>0$.
An edge-labeled 1-WL refinement distinguishes these graphs after one
iteration, and a message-passing GNN that conditions aggregation on $e(u,v)$
can simulate this refinement.

Hence, non-local edge attributes can strictly increase expressive power
beyond 1-WL.
\end{proof}

\subsection{Proof of Theorem~\ref{thm:pe-expressivity}}
\label{app:proof-pe-expressivity}

\begin{proof}
Let $G$ be a graph with nontrivial automorphisms, such that 1-WL assigns the
same color to all nodes.
Assign to each node $v$ a positional encoding $p_v\in\mathbb{R}^k$ such that
\[
\sum_{u\in S}p_u \neq \sum_{u\in T}p_u
\quad\text{for all distinct multisets } S\neq T.
\]
This condition holds with probability one if the $p_v$ are drawn from a
continuous distribution, or when Laplacian eigenvectors are in general
position (i.e., do not satisfy accidental linear dependencies).

Define
\[
s(v) := \sum_{u\in\mathcal{N}(v)} p_u.
\]
Under the stated condition, the mapping $v\mapsto (p_v,s(v))$ uniquely
determines the neighborhood multiset of $v$.
A single message-passing layer can compute $s(v)$, and an MLP can map
$(p_v,s(v))$ to unique node representations.

Thus, the GNN distinguishes nodes and hence graphs that 1-WL cannot
distinguish
\end{proof}

\subsection{Proof of Proposition~\ref{prop:finite-termination}}
\label{app:proof-finite-termination}

\begin{proof}
Since $V$ is fixed, the space of admissible edge sets is finite.
If $\mathcal{L}(E)$ strictly decreases whenever rewiring changes the edge set,
then the sequence $\{\mathcal{L}(E_t)\}$ is strictly decreasing and cannot
cycle. Hence the process must reach a fixed point $E^\star$ after finitely
many steps, i.e., $E_{t+1}=E_t=E^\star$.
\end{proof}

\subsection{Proof of Theorem~\ref{thm:local-stabilization}}
\label{app:proof-local-stabilization}

\begin{proof}
We show that both the added-edge set and the pruned-edge set are locally
unchanged near $E^\star$.

\textbf{(LapPE stability).}
Perturbations of the normalized Laplacian in
operator norm induce perturbations of the Laplacian eigenvectors of order
$O(\|\Delta L\|/\mathrm{gap})$ on the retained subspace. Under the eigengap
assumption, LapPE vary continuously (indeed Lipschitzly) for $E$ close enough
to $E^\star$.

\textbf{(Stability of $k$NN additions).}
By the $k$NN margin assumption, sufficiently small perturbations of LapPE do
not change each node's set of $k$ nearest neighbors. Hence the set of added
edges is identical to that at $E^\star$ for all $E$ in a small neighborhood.

\textbf{(Stability of pruning).}
By the pruning-score margin and the score stability assumption, the ordering
of positive-curvature edge scores near the pruning threshold is unchanged for
all $E$ close enough to $E^\star$. Therefore, the top
$\lceil\rho |E^+|\rceil$ edges selected for removal are the same as at
$E^\star$.

Combining these two facts, the rewiring step performs exactly the same edge
deletions and additions as at $E^\star$, implying
$\mathcal{R}(E)=\mathcal{R}(E^\star)=E^\star$.
\end{proof}

\subsection{Proof of Lemma~\ref{lem:operator-perturbation}}
\label{app:proof-operator-perturbation}

\begin{proof}
Consider a single edge insertion or deletion.
Let $A'$ and $D'$ denote the resulting adjacency and degree matrices.
We write
\begin{align}
\tilde A(E')-\tilde A(E)
&=
(D'^{-1/2}-D^{-1/2})(A+I)D^{-1/2} \nonumber\\
&\quad+\; D'^{-1/2}(A'-A)D^{-1/2} \nonumber\\
&\quad+\; D'^{-1/2}(A'+I)(D'^{-1/2}-D^{-1/2}).
\label{eq:deltaA-decomp}
\end{align}

The adjacency difference $A'-A$ is a rank-two matrix with operator norm at
most $1$, and the (self-loop augmented) degree of any node changes by at most
$1$.
For diagonal degree matrices, applying the mean value theorem to
$f(x)=x^{-1/2}$ yields
\[
\|D'^{-1/2}-D^{-1/2}\|_2
\;\le\;
\frac{1}{2\,\tilde d_{\min}^{3/2}}.
\]
Moreover, using the factorization
$(A+I)D^{-1/2}=D^{1/2}\tilde A(E)$, we obtain
\[
\|(A+I)D^{-1/2}\|_2
\;\le\;
\sqrt{\tilde d_{\max}}\;\|\tilde A(E)\|_2
\;\le\;
\sqrt{\tilde d_{\max}},
\]
where $\|\tilde A(E)\|_2\le 1$ for the normalized operator.
Under the bounded degree ratio assumption,
$\sqrt{\tilde d_{\max}} \le \sqrt{c\,\tilde d_{\min}}$.

Using submultiplicativity of the operator norm and
$\|D^{-1/2}\|_2,\|D'^{-1/2}\|_2\le 1/\sqrt{\tilde d_{\min}}$, we obtain
\[
\|D'^{-1/2}(A'-A)D^{-1/2}\|_2
\;\le\;
\frac{1}{\tilde d_{\min}},
\]
and
\[
\|(D'^{-1/2}-D^{-1/2})(A+I)D^{-1/2}\|_2
\;\le\;
\frac{\sqrt{c}}{2\,\tilde d_{\min}}.
\]
The third term in~\eqref{eq:deltaA-decomp} is bounded analogously.
Hence, a single edge edit perturbs $\tilde A$ by at most
$C/\tilde d_{\min}$ for a constant $C$ depending only on $c$.
By subadditivity of the operator norm, $K$ such edits yield
\[
\|\tilde A(E')-\tilde A(E)\|_2
\;\le\;
C\,\frac{K}{\tilde d_{\min}}.
\]
\end{proof}

\subsection{Proof of Corollary~\ref{cor:lipschitz-stability}}
\label{app:proof-lipschitz-stability}

\begin{proof}
By Lemma~\ref{lem:operator-perturbation},
$\|\tilde A(E')-\tilde A(E)\|_2 \le C K/\tilde d_{\min}$.
Each layer map is Lipschitz with constant at most
$\|\tilde A(E)\|_2\|W_\ell\|_2$,
and $\|\tilde A(E)\|_2\le 1$ for the normalized operator.
Composing the perturbation through $T$ layers yields the stated bound.
\end{proof}

\subsection{Proof of Proposition~\ref{prop:when-geometry-helps}}
\label{app:proof-when-geometry-helps}

\begin{proof}
We prove the two parts separately.

\textbf{Necessity.}
If $I(Y;E\mid \tilde X)=0$, then $Y$ and $E$ are conditionally independent
given $\tilde X$.
Hence
\[
p(Y\mid \tilde X,E)=p(Y\mid \tilde X)\quad\text{a.s.}
\]
The Bayes-optimal predictor using $(\tilde X,E)$ therefore coincides with the
Bayes-optimal predictor using $\tilde X$ alone, implying
\[
R^\ast(\tilde X,E)=R^\ast(\tilde X).
\]

\textbf{Sufficiency.}
If $I(Y;E\mid \tilde X)>0$, then
\[
H(Y\mid \tilde X,E)<H(Y\mid \tilde X),
\]
so the Bayes-optimal edge-aware predictor achieves strictly lower risk than
the Bayes-optimal $\tilde X$-only predictor.
\end{proof}

\section{Related Work}
\label{app:related}
\paragraph{Homophily, heterophily, and label informativeness.}
The effectiveness of message-passing GNNs has traditionally been explained
through \emph{homophily}, the tendency of neighboring nodes to share labels
\cite{kipf2017semi}.
However, recent work has shown that homophily alone is insufficient: graphs
with similar homophily levels can exhibit vastly different GNN behavior, and
some heterophilous graphs remain amenable to standard message passing
\cite{zhu2020beyond,ma2022is}.
To address this limitation, adjusted homophily (assortativity) and
\emph{label informativeness} (LI) were proposed as more principled descriptors
of graph structure \cite{platonov2023characterizing}.
LI measures how informative a neighbor’s label is about a node’s label and
has been shown to correlate more strongly with GNN performance than raw
homophily.
Our work builds on this line of research by providing an information-theoretic
criterion for when graph structure can improve prediction, and by analyzing
how geometric mechanisms interact with label informativeness.

\paragraph{Curvature and structural encodings.}
Discrete Ricci curvature notions, including Forman and Ollivier curvature,
have been used to characterize local graph geometry
\cite{forman2003bochner,ollivier2009ricci,sreejith2016forman}.
Recent work introduced curvature-based structural encodings and local
curvature profiles as node features for GNNs, reporting empirical gains
especially on heterophilous benchmarks \cite{fesser2024effective}.
These methods are typically motivated heuristically.
In contrast, we show that degree-based Forman curvature does not increase
expressivity beyond 1-WL, but can improve performance by reshaping diffusion
and reducing harmful cross-class mixing.

\paragraph{Positional encodings.}
Positional encodings have been widely studied as a means of injecting global
structure into GNNs \cite{dwivedi2020generalization, dwivedi2022lspe}.
Laplacian eigenvectors and related spectral features have been shown to
mitigate over-smoothing and to increase expressive power by breaking graph
symmetries.
Our analysis complements this work by formalizing how Laplacian positional
encodings counteract the low-pass bias of Laplacian smoothing and interact
with geometry-aware message passing.

\paragraph{Graph rewiring.}
Graph rewiring and topology augmentation have been proposed to alleviate
information flow bottlenecks such as over-squashing in deep GNNs
\cite{topping2022understanding}.
Existing approaches include heuristic edge pruning, curvature-guided rewiring, and learned topology modification. Although effective in practice, most rewiring methods come with limited
algorithmic guaranties.
We contribute to this literature by analyzing a simple curvature-guided
rewiring process, establishing finite termination, local stabilization, and
Lipschitz stability under bounded edge edits.

\paragraph{Expressivity of GNNs.}
The expressive power of message-passing GNNs is known to be bounded by the
one-dimensional Weisfeiler--Lehman (1-WL) test \cite{xu2019powerful}.
Recent work has shown that augmenting GNNs with random node features 
can enable them to distinguish some graphs that 1-WL cannot \cite{sato2021random}.
Our results refine this understanding by separating local, degree-based
geometric signals from non-local edge attributes and positional encodings,
and by identifying which geometric augmentations can and cannot exceed the
1-WL expressivity ceiling.

Unlike prior work that studies curvature, positional encodings, or rewiring in
isolation, we present a unified theoretical framework that connects these
mechanisms through label informativeness and conditional edge--label
information, clarifying when geometry-aware GNNs should be expected to help.

\section{Experimental Details}
\label{app:experiments}

\subsection{Architecture}
\label{app:architecture}
\textbf{ASE base} is a geometry-augmented but \emph{static} model that
concatenates raw node features with local curvature profiles (LCP) and
Laplacian positional encodings (LapPE), without modifying graph structure
or using edge-level curvature during message passing.
\textbf{ASEHybrid} extends ASE base by incorporating curvature-aware
attention and optional curvature-guided rewiring.

Concretely, ASEHybrid follows the pipeline:
LayerNorm $\rightarrow$ GCNConv $\rightarrow$ ELU $\rightarrow$
PairNorm~\cite{zhao2020pairnorm} $\rightarrow$
GATv2~\cite{brody2022attentive} ($\texttt{edge\_dim}=1$) $\rightarrow$ Linear.
Input features concatenate raw node attributes with
LCP (five statistics: min/max/mean/std/median of incident Forman
curvatures) and Laplacian positional encodings (LapPE; $K=8$
eigenvectors)~\cite{sato2021random}.
All graphs are symmetrized, self-loops removed, and edges coalesced.
\subsection{Training Protocol}
\label{app:training}
All models are trained using AdamW
(learning rate $0.003$--$0.005$ for ASEHybrid, $0.01$ for GCN;
weight decay $5\times10^{-4}$),
for $700$--$800$ epochs with early stopping
(patience $120$--$150$).
Unless stated otherwise, results are reported as mean $\pm$ standard
deviation over \textbf{10 random seeds} for datasets with a fixed split
(Chameleon, Squirrel), and over the \textbf{dataset-provided splits}
(10 in PyTorch Geometric) for WebKB--Texas and
HeterophilousGraphDataset benchmarks (Minesweeper, Tolokers).

\subsection{Minesweeper: High-Baseline Regime}
\label{app:minesweeper}
Minesweeper is a synthetic benchmark where node features alone are
highly predictive and label informativeness of edges is minimal.
The theory predicts that geometry-aware mechanisms should not improve
performance in this regime.

As expected, all methods perform nearly identically
(see Table~\ref{tab:main-results}).
This serves as a necessary sanity check, empirically
confirming the theoretical claim that when
$I(Y;E\mid \tilde X)=0$, no geometric augmentation can
improve Bayes-optimal prediction.

\subsection{Curvature-Guided Rewiring Algorithm}
\label{app:rewiring-algorithm}

Let $G_t=(V,E_t)$ denote the graph at rewiring step $t$, with $m_t:=|E_t|$.
Given node representations (or positional encodings) and associated geometric
signals, a single rewiring step applies:

\begin{enumerate}
    \item \textbf{Score and prune edges.}
    Each edge $(u,v)\in E_t$ is assigned a score $s_t(u,v)$; in our instantiation,
    $s_t(u,v)$ is the Forman curvature of the edge computed on $G_t$.

     Pruning is restricted to edges with
    \emph{positive} score, which we denote by
    \[
    E_t^+ := \{e\in E_t : s_t(e) > 0\}.
    \]
    Let $(e_1,\dots,e_{|E_t^+|})$ be an ordering of $E_t^+$ such that
    \[
    s_t(e_1)\ge s_t(e_2)\ge \cdots \ge s_t(e_{|E_t^+|}).
    \]
    For a fixed $\rho\in(0,1)$, remove the top
    $\lceil \rho |E_t^+|\rceil$ edges
    $\{e_1,\dots,e_{\lceil\rho |E_t^+|\rceil}\}$.
    If $E_t^+=\emptyset$, no edges are pruned.

    \item \textbf{Add $k$NN edges in LapPE space.}
    Compute Laplacian positional encodings $P_t=[\phi_2,\dots,\phi_{K+1}]$ on
    $G_t$. For each node $u$, add edges to its $k$ nearest neighbors under the
    Euclidean metric in LapPE space (excluding existing neighbors and $u$
    itself).

    \item \textbf{Update.}
    Recompute the geometric scores on the updated graph to obtain $G_{t+1}$.
\end{enumerate}

This defines a (possibly stochastic) rewiring operator
\[
\mathcal{R}:E_t\mapsto E_{t+1}.
\]

\subsection{Implications for Geometry-Aware GNNs}
\label{app:implications}
\paragraph{Why curvature and rewiring help.}
Curvature-based encodings and rewiring modify how information from $E$ is
propagated and emphasized.
When $I(Y;E\mid \tilde X)>0$, these mechanisms can amplify informative edges
and suppress misleading ones, effectively increasing the usable signal
extracted from the graph.

\paragraph{Why geometry sometimes fails.}
When $I(Y;E\mid \tilde X)=0$, no manipulation of $E$—including curvature
weighting, positional encodings derived from $E$, or rewiring can improve
prediction in principle.
In such regimes, empirical gains are not expected, and $\tilde X$-only
baselines are optimal.

\paragraph{Relation to label informativeness.}
Label informativeness (LI) measures mutual information between labels at the
endpoints of a random edge.
While LI is not identical to $I(Y;E\mid \tilde X)$, it provides a practical
proxy: high LI suggests that edges are informative about labels, whereas low
LI indicates that edges are uninformative given node-side information.
Our theory (Proposition~\ref{prop:when-geometry-helps} explains why geometry-aware GNNs tend to succeed precisely on
datasets with high LI.

\paragraph{Summary.}
The usefulness of geometry-aware message passing is not determined by
homophily alone, nor by architectural sophistication.
Rather, it is fundamentally governed by whether graph structure contains
label-relevant information beyond the node-side information $\tilde X$
already available to the predictor.


\end{document}